\RequirePackage{fix-cm}
\RequirePackage{amsmath}
\documentclass[smallcondensed]{svjour3}       
\smartqed  

\usepackage{graphicx}
\usepackage{mathptmx}      
\usepackage{newtxtext,newtxmath}
\usepackage{amsfonts}
\usepackage{tikz}
\usepackage{pgfplots}
\usepgfplotslibrary{groupplots}
\usetikzlibrary{matrix}
\usepackage{soul}
\usepackage{subfigure}
\usepackage[normalem]{ulem}
\usepackage{wrapfig,lipsum,booktabs}
\usepackage[round]{natbib}
\usepackage{url}
\usepackage{lscape}
\usepackage{hyperref}
\pgfplotsset{
    ComplexFig/.style={
      width=0.52\textwidth,
      ymin=0,
      enlargelimits = false,
      x tick label style={
        /pgf/number format/.cd,
            fixed,
            precision=2,
        /tikz/.cd
      },
      scaled x ticks=false,
      every axis plot/.append style={line width=1.0pt},
    },
    ComplexTest/.style={color=black,mark=none},
    ComplexDisagree/.style={color=black, dashed,mark=none},
    ComplexPB/.style={color=blue,mark=none},
    ComplexC1/.style={color=green,mark=none},
    ComplexC2/.style={color=red,mark=none},
    ComplexSH/.style={color=yellow,mark=none},
}

\newcommand{\tsize}{n}

\newcommand{\set}{S}
\newcommand{\trainset}{T}
\newcommand{\valset}{T_{\text{\normalfont{val}}}}
\newcommand{\testset}{T_{\text{\normalfont{ext}}}}

\newcommand{\oobvalset}{\bar T}
\newcommand{\Xspace}{\mathcal{X}}
\newcommand{\Yspace}{\mathcal{Y}}
\newcommand{\Hspace}{\mathcal{H}}

\newcommand{\prior}{\pi}
\newcommand{\posterior}{\rho}
\newcommand{\lr}[1]{\left(#1\right)}
\newcommand{\lrs}[1]{\left[#1\right]}

\newcommand{\E}{\mathbb E}

\renewcommand{\P}{\mathbb P}
\newcommand{\loss}{\ell}
\newcommand{\margin}{\mathcal{M}}
\newcommand{\uniform}{u}
\newcommand{\PB}{PBkl}
\newcommand{\C}{\ensuremath{C}}
\newcommand{\sinhyp}{SH}
\newcommand{\sh}{\sinhyp}
\newcommand{\shbound}{\sh-bound}
\newcommand{\datadist}{D}
\newcommand{\pbbound}{\PB-bound}
\newcommand{\lambdabound}{$\lambda$-bound}
\newcommand{\cbound}{$\C$-bound}
\newcommand{\disagree}{d_{\posterior}}
\newcommand{\disagreeemp}{\hat{d}_{\posterior}}
\newcommand{\jerr}{e}
\newcommand{\jerremp}{\hat{e}_{\posterior}}
\newcommand{\indicator}[1]{I\lr{#1}}
\newcommand{\cboundempA}{$\C1$-bound}
\newcommand{\cboundempB}{$\C2$-bound}
\newcommand{\cboundempC}{$\C3$-bound}
\newcommand{\hypothesis}{h}
\newcommand{\hyp}{\hypothesis}
\newcommand{\gvoter}{{h_G}}
\newcommand{\gloss}{L^{\text{\normalfont{Gibbs}}}(\gvoter)}
\newcommand{\glossemp}[1]{\hat L^{\text{\normalfont{Gibbs}}}(\gvoter,#1)}
\newcommand{\glossoob}[1]{\hat L_{\text{OOB}}^{\text{\normalfont{Gibbs}}}(\gvoter,#1)}
\newcommand{\mvoter}{{h_M}}
\newcommand{\mloss}{L^{\text{\normalfont{\normalfont{MV}}}}(\mvoter)}
\newcommand{\mlossemp}[1]{\hat L^{\text{\normalfont{MV}}}(\mvoter,#1)}
\newcommand{\mlossoob}[1]{\hat L_{\text{OOB}}^{\text{\normalfont{MV}}}(\mvoter,#1)}
\newcommand{\hcount}{m}
\newcommand{\dataset}[1]{\textsc{#1}}
\newcommand{\datasetsize}{N}

\DeclareMathOperator{\KL}{KL}
\DeclareMathOperator{\kl}{kl}

\newcommand{\ci}[1]{\textcolor{blue}{#1}}
\definecolor{cbagg}{RGB}{198, 198, 198}
\definecolor{cbsin}{RGB}{200,127,0}
\definecolor{ctest}{RGB}{255, 205, 119}
 \journalname{Machine Learning}
\begin{document}

\title{On PAC-Bayesian Bounds for Random Forests
  \thanks{The  work  is  supported  by  the  Innovation  Fund  Denmark through the Danish Center for Big Data Analytics Driven Innovation (DABAI) project.}
}


\author{Stephan S.~Lorenzen \and
  Christian~Igel      \and
  Yevgeny~Seldin
}

\authorrunning{S.~Lorenzen \and C.~Igel \and Y.~Seldin} 

\institute{S.~Lorenzen \and C.~Igel \and Y.~Seldin \at
  Department of Computer Science, University of Copenhagen \\
  \email{\{lorenzen,igel,seldin\}@di.ku.dk}
}

\date{Received: date / Accepted: date}

\maketitle

\begin{abstract}
  Existing guarantees in terms of rigorous upper bounds on the generalization error for the original random forest algorithm, one of the most frequently used machine learning methods, are unsatisfying.  We discuss and evaluate various PAC-Bayesian approaches to derive such bounds. The bounds do not require additional hold-out data, because the out-of-bag samples from the bagging in the training process can be exploited.

  A random forest predicts by taking a majority vote of an ensemble of decision trees.  The first approach is to bound the error of the vote by twice the error of the corresponding Gibbs classifier (classifying with a single member of the ensemble selected at random). However, this approach does not take into account the effect of averaging out of errors of individual classifiers when taking the majority vote. This effect provides a significant boost in performance when the errors are independent or negatively correlated, but when the correlations are strong the advantage from taking the majority vote is small.  The second approach based on PAC-Bayesian \cbound s takes dependencies between ensemble members into account, but it requires estimating correlations between the errors of the individual classifiers. When the correlations are high or the estimation is poor, the bounds degrade.

  In our experiments, we compute generalization bounds for random
  forests on various benchmark data sets.  Because the individual
  decision trees already perform well, their predictions are highly
  correlated and the \cbound s do not lead to satisfactory
  results. For the same reason, the bounds based on the analysis of
  Gibbs classifiers are typically superior and often reasonably tight.
  Bounds based on a validation set coming at the cost of a smaller training
  set   gave better performance guarantees,
  but worse performance in most experiments.

\end{abstract}

\section{Introduction}
\label{section:intro}
A \textit{random forest} is one of the most successful machine learning algorithms \citep{Breiman2001}.  It is easy to use and to parallelize and often achieves high accuracies in practice \citep{Delgado2014}.  In a survey on the machine learning competition website kaggle.com\footnote{http://www.kaggle.com/surveys/2017}, 46\% of 16.000 surveyed users claimed to use the algorithm in their daily work.
{A random forest for classification predicts  based on the (possibly weighted) \emph{majority vote}} of a set (an \emph{ensemble}) of weaker classifiers, concretely \emph{decision trees}.
The model was first presented by \cite{Breiman2001}, who provides an initial analysis and some theoretical bounds, showing that the strength of the random forest depends on the strength of individual trees and their correlation. 
Despite its popularity in practice, the algorithm is still not well understood theoretically \citep{Arlot2014,Biau2012,Denil2014}, the main reason being that the model is difficult to analyse because of the dependencies between the induced partitions of the input space and the predictions within the partitions \citep{Arlot2014}. The conceptually simpler \emph{purely random forests} \citep{Breiman2002} avoids these dependencies by creating a random partitioning independent of the training data. This is done by selecting features and splits at random. \cite{Biau2008} show the purely random forests to be \emph{consistent}
under some assumptions on the distribution of the input variables.
Several modification of the random forest have been introduced in the literature, most of them being in between the standard random forest and the purely random forest in the sense that extra randomness is added to get independent partitions \citep{Geurts2006,Genuer2010}. For instance, \cite{Wang2016} introduce the \emph{Bernoulli random forests}, which relies on Bernoulli random variables for randomly choosing the strategy for partitioning the input space, and prove this model to be consistent. Likewise, \cite{Denil2014} give a variant based on sampling of predictions in a partition for determining best splits, and prove this variant to be consistent.
Theoretical bounds on the expected loss have been considered by \citet{Genuer2010} in the case of regression tasks when the input space is one-dimensional. \citet{Arlot2014} consider the generalization error for the purely random forest in relation to the number of trees.   
All these have nice analytical properties, but these come at the expense of degradation in empirical performance compared to the standard random forest. Accordingly, the original random forest still remains the best choice in practice \citep{Wang2016}, despite the lack of strong theoretical guarantees.

{\sloppy
  This study considers the application of theoretical bounds based on
  PAC-Bayesian analysis to the standard random forest as given by
  \cite{Breiman2001}. Here PAC stands for the Probably Approximately
  Correct frequentist learning model \citep{Valiant1984}. PAC-Bayesian approaches 
are usually used for analysing the expected loss of \textit{Gibbs classifiers}. Gibbs classifiers are randomized  classifiers that make predictions by applying a hypothesis drawn from a hypothesis class $\Hspace$ according to some distribution $\posterior$ on $\Hspace$ \citep{McAllester1998,Seeger2002,Thiemann2017}. While generalization bounds for Gibbs classifier may at first not seem directly applicable to majority vote classifiers, they are in fact closely related. It can be shown that the loss of a
$\posterior$-weighted majority vote classifier is at most twice that of the associated Gibbs classifier,
meaning that any bound for a Gibbs classifier leads to a bound for the majority vote \citep{Germain2015}.
However, such adaptation of the bounds for Gibbs classifiers typically provides relatively weak bounds for majority vote classifiers, because the bounds for Gibbs classifiers do not take into account dependencies between individual classifiers.
One of the main reasons for the good performance of  majority vote classifiers is that 
when the errors of individual classifiers are independent they tend to average out \citep{Breiman2001}. Therefore, the majority vote may perform very well even when the individual classifiers are weak (i.e., only slightly better than random guessing). In this case, application of PAC-Bayesian bounds for the Gibbs classifier to the majority vote yields suboptimal results. 
}

This has motivated the development of PAC-Bayesian bounds designed specifically for averaging and majority vote classifiers \citep{Germain2015,McAllester1999,ONETO201821}. One such bound is the \emph{\cbound}, given by \citet{Germain2015}, which is based on the \emph{margin} of the classifier.
In contrast to the bounds for Gibbs classifiers, the \cbound\ takes the correlations between the individual classifiers into account and could potentially yield tighter bounds in the case described above. However, in the case with strong individual classifiers and high correlation (as is the case for random forests), the \cbound\ deteriorates \citep{Germain2015} -- in contrast to the 
Gibbs classifier bounds.



In this study,
several of the above mentioned bounds are applied to the standard random forest setting, where trees are trained using \emph{bagging}, that is, using different random subsets of the training data \citep{Breiman2001,Breiman1996b}. Since validation sets for individual trees are constructed as part of the training procedure when using bagging, the theoretical bounds come ``for free'' in the sense that no separate data needs to be reserved for evaluation.
We compare the quality of bounds obtained in this setting with bounds obtained by leaving out a validation set for evaluation. We also consider optimization of the weighting of the voters by minimization of the theoretical bounds \citep{Thiemann2017,Germain2015}.


\section{Background}
\label{section:background}
We consider supervised learning. Let $\set = \{(X_1,Y_1), \ldots, (X_\tsize, Y_\tsize)\}$ be an independent identically distributed sample from $\Xspace \times \Yspace$, drawn according to an unknown distribution $\datadist$. A hypothesis is a function $\hypothesis : \Xspace \to \Yspace$, and $\Hspace$ denotes a space of hypotheses. We evaluate a hypothesis $h$ by  a bounded loss function $\loss : \Yspace^2 \to [0,1]$.
The expected loss of $\hyp$ is denoted by $L(\hypothesis) = \E_{(X,Y)\sim D}\left[\loss(\hypothesis(X),Y)\right]$ and the empirical loss of $\hyp$ on a sample $\set$ is denoted by $\hat L(\hypothesis, \set) = \frac{1}{n}\sum^{n}_{i=1} \loss(\hypothesis(X_i),Y_i)$.
In this study, we focus on classification.
%
Given a set of hypotheses $\Hspace$ and a distribution $\posterior$ on $\Hspace$, the \emph{Gibbs classifier} $\gvoter$ is a stochastic classifier, which for each input $X$ randomly draws a hypothesis $\hypothesis\in\Hspace$ according to $\posterior$ and predicts $\hypothesis(X)$ \citep{Seeger2002}. The expected loss of the Gibbs classifier is given by $\gloss = \E_{\hypothesis\sim\posterior}\lrs{L(\hypothesis)}$, and the empirical loss of $\gvoter$ on a sample $\set$ is given by $\glossemp{\set} = \E_{\hypothesis\sim\posterior}\lrs{\hat L(\hypothesis, \set)}$.

Closely related to the random Gibbs classifier are \emph{aggregate classifiers}, whose predictions are based on weighted aggregates over $\Hspace$. The $\posterior$-weighted \emph{majority vote} $\mvoter$ predicts $\mvoter(X) = \operatorname{argmax}_{Y\in\Yspace}
\sum_{\hypothesis\in\Hspace \land \hypothesis(X)=Y}\rho(h)$. When discussing majority vote classifiers, it is convenient to define the \emph{margin} realised on a pattern $(X,Y)$ \citep{Breiman2001}:
\begin{equation}
  \margin_\posterior(X,Y) = \P_{\hyp\sim\posterior}\lrs{\hyp(X)=Y}-\max_{j\neq Y}\P_{\hyp\sim\posterior}\lrs{\hyp(X)=j},
  \label{eq:margin}
\end{equation}
and the expected value of the margin $\margin_\posterior = \E_{(X,Y)\sim\datadist}\lrs{\margin_\posterior\lr{X,Y}}$. Note, that a large margin indicates a strong classifier. The expected loss of $\mvoter$ is then given by $\mloss = \P_{(X,Y)\sim\datadist}\lrs{\margin_\posterior(X,Y) \leq 0}$, and the empirical loss $\mlossemp{\set} = \P_{(X,Y)\sim\set}\lrs{\margin_\posterior(X,Y) \leq 0}$, where we use $(X,Y)\sim\set$ to denote a uniform distribution over the sample.

{The Kullback-Leibler divergence between two distributions $\prior$ and $\posterior$ is denoted by $\KL(\posterior\|\prior) $ and between Bernoulli distributions with biases $p$ and $q$ by $\kl(p\|q)$.
  Furthermore, let $\E_\datadist[\cdot]$ denote $\E_{(X,Y)\sim\datadist}[\cdot]$ and $\E_{\rho}[\cdot]$ denote $\E_{\hypothesis\sim\rho}[\cdot]$. Finally, $\uniform$ denotes the uniform distribution.}

\subsection{Random Forests.}
Originally described by \citet{Breiman2001}, the random forest is a majority vote classifier, where individual voters are decision trees. In the standard random forest setting, every voter has equal weight (i.e., $\rho = \uniform$).
Let $\trainset \subset \Xspace \times \Yspace$ denote training patterns drawn according to $\datadist$. A random forest is constructed by independently constructing \emph{decision trees} $\hyp_1, \hyp_2, ..., \hyp_\hcount$ \citep{Hastie2009}, where each $\hypothesis_i$ is trained on $\trainset_i\subseteq\trainset$. A tree is constructed recursively, starting at the root. At each internal node, a threshold $\theta$ and a feature $j$ are chosen, and the data set $\trainset'$ corresponding to the current node is then split into $\{X\, |\, X_j \leq \theta\}$ and $\{ X\, |\, X_j > \theta \}$.
$\theta$ and $j$ are chosen according to some splitting criterion, usually with the goal of maximizing the \emph{information gain} for each split, making the new subsets more homogeneous \citep{Hastie2009}. Splitting is stopped when a node is completely \emph{pure} (only one class is present) or by some other stopping criterion  (e.g., maximum tree depth).
The tree construction is randomized \citep{Breiman1996b,Breiman2001}. First, the selection of data sets $\trainset_i$ for training individual trees is randomized. They are generated by the bagging procedure described below. Second, only a random subset of all features is considered when splitting at each node \citep{Breiman2001,Hastie2009}.

Bagging is a general technique used for aggregated predictors \citep{Breiman1996b}, which generates the training sets $\trainset_i$ for the individual predictors by sampling  $|\trainset|$ points from $\trainset$ with replacement. The individual training sets $\trainset_1,\trainset_2,...$ are known as \emph{bootstrap samples}.
Because of sampling with replacement, not all patterns of $\trainset$ are expected to be in each $\trainset_i$. Let $\oobvalset_i = \trainset \setminus \trainset_i$ denote the patterns not sampled for $\trainset_i$. $\oobvalset_i$ can now be used to give an unbiased estimate of the individual classifier $\hyp_i$. The expected number of unique patterns in $\trainset_i$ is approximately $\lr{1-\frac{1}{e}}|\trainset| \simeq 0.632|\trainset|$,
leaving us with slightly more than one third of the training patterns for the validation sets \citep{Breiman1996b}.

Bagging also allows us to compute \emph{out-of-bag (OOB)} estimates. For each training pattern $(X,Y) \in \trainset$, a majority vote prediction is computed over all voters $\hyp_i$ with $(X,Y)\not\in \trainset_i$. The empirical loss computed over these predictions is known as the OOB estimate, which we denote by $\mlossoob{\trainset}$. 
Empirical studies have shown that the OOB estimate on the training data is a good estimator of the generalization error  \citep{Breiman1996a}.

Furthermore, the sets
$\oobvalset_1, \oobvalset_2, ..., \oobvalset_\hcount$
can be used to compute the empirical error of the associated $\posterior$-weighted Gibbs classifier $\gvoter$ by
\begin{equation*}
  \glossoob{\trainset} = \E_\posterior\lrs{\frac{1}{|\oobvalset_i|}\sum_{(X,Y)\in \oobvalset_i} \loss\lr{\hyp_i(X),Y}},
\end{equation*}
and by considering $\oobvalset_i \cap \oobvalset_j$,
we can also estimate the correlation between trees $\hyp_i$ and $\hyp_j$,
an important ingredient in  bounds for majority vote classifiers.

\section{PAC-Bayesian Bounds for Majority Vote Classifiers}
\label{section:bounds}

We now give an overview of the PAC-Bayesian bounds we apply to bound the expected loss of random forests.
PAC-Bayesian bounds have a form of a trade-off between $\rho$-weighted empirical loss of hypotheses in ${\cal H}$ and the complexity of $\rho$, which is measured by its Kullback-Leibler divergence from a prior distribution $\pi$. The prior must be selected before the data is observed and can be used to incorporate domain knowledge, while the posterior $\rho$ can be chosen based on the data.

\subsection{PAC-Bayesian Bounds for Gibbs Classifiers.}
The $\posterior$-weighted majority vote classifier $\mvoter$ is
closely related to the $\posterior$-parameterized Gibbs classifier
$\gvoter$. Whenever the majority vote makes a mistake, it means that
more than a $\posterior$-weighted half of the voters make a
mistake. Thus,
{the expected loss of the majority vote classifier $\mloss$ is at
  most twice the
expected loss of the Gibbs classifier $\gloss$, }
\begin{equation}
  \mloss \leq 2 \gloss \label{eq:factor2}
\end{equation}
\citep{McAllester2003,Langford2002,Germain2015}. Therefore, any bound
on {$\gloss$  provides a bound on the corresponding $\mloss$.} 
We consider the following inequality originally due to \citet{Seeger2002}, which we refer to as the \pbbound\ {(\textbf{P}AC-\textbf{B}ayesian \textbf{kl})}:
\begin{theorem}[\pbbound, \citealp{Seeger2002}]For any probability distribution $\pi$ over ${\cal H}$ that is independent of $S$ and any $\delta \in (0,1)$, with probability at least $1-\delta$ over a random draw of a sample $S$, for all distributions $\rho$ over ${\cal H}$ simultaneously:
  \begin{equation*}
    \label{eq:PBkl}
    \kl\lr{\glossemp{\set}\middle\|\gloss} \leq \frac{\KL(\rho\|\pi) + \ln \frac{2 \sqrt \tsize}{\delta}}{\tsize}.
  \end{equation*}
  \label{thm:PBkl}
\end{theorem} %
{A slightly tighter bound can be obtained by using
\begin{equation}
  \xi(\tsize) = \sum^\tsize_{k=0} \binom{\tsize}{k} \lr{\frac{k}{\tsize}}^k \lr{1-\frac{k}{\tsize}}^{\tsize-k}
  \label{eq:xi-def}
\end{equation}
instead of  $2\sqrt{\tsize}$ in the bound above, because we have $\sqrt{\tsize} \leq \xi(\tsize) \leq 2\sqrt{\tsize}$ \citep{Mau04}.}

In order to use Theorem~\ref{thm:PBkl} in the bagging setting, we need to make a small adjustment.
The empirical Gibbs loss $\glossemp{\trainset}$ is computed using $\oobvalset_1, \oobvalset_2, ...$, and since these sets have different sizes, in order to apply the \pbbound, we use
$\tsize = \min_i\lr{|\oobvalset_i|}$.
That this is a valid strategy can easily be seen by going through the proof of Theorem~\ref{thm:PBkl}, see \citet{Thiemann2017}.

Theorem~\ref{thm:PBkl} may also be applied to the final majority vote
classifier $\gvoter$ if a separate validation set is left out.
  In this case, $|\Hspace| = 1$, $\KL(\rho\|\pi) = 0$, and $\hat
    L^{\text{\normalfont{Gibbs}}}(\mvoter,\set) = \mlossemp{\set}$.
A separate validation set implies
that the data available at training time has to be split, and, 
therefore, the actual training set gets smaller. While  $\hat
    L^{\text{\normalfont{Gibbs}}}(\mvoter,\set)$ 
  may be larger due to the smaller training data set size,  the
  bound does no longer suffer from the factor 2 in 
\eqref{eq:factor2}. We will consider this way of bounding $\mloss$
as an additional baseline denoted as \textit{\shbound} {(\textbf{S}ingle \textbf{H}ypothesis)}. Note that this bound requires the separate validation set and, thus, cannot be applied in the bagging setting.

\subsection{PAC-Bayesian Bounds for Majority Vote Classifiers.}
The \pbbound\ provides a tight bound for the Gibbs classifier, but the associated bound for the majority vote classifier may be loose. This is because the bound for the Gibbs classifier does not take correlation between individual classifiers into account.
The individual classifiers may be weak (i.e., $L(h_i)$ close to $\frac{1}{2}$) leading to a weak Gibbs classifier, but if the correlations between the classifiers are low, the errors tend to cancel out when voting, giving a stronger majority vote classifier
\citep{Germain2015}. The generalization bounds for Gibbs classifiers do not capture this, as they depend only on the strength of the individual classifiers. 
In order to get stronger generalization guarantees for majority vote classifiers, we need bounds that incorporate information about the correlations between errors of classifiers as well, as already pointed out by \citet{Breiman2001}.

\cite{Germain2015} propose to use the \cbound\ for this purpose, which is based on the margin of the majority vote classifier. They consider only the case where the output space is binary, $\Yspace = \{-1,1\}$, and \eqref{eq:margin} becomes 
$\margin_\posterior(X,Y) = Y\lr{\sum_{\hyp\in\Hspace} \posterior(\hyp) \hyp(X)}$.
With the first moment $\margin_\posterior = \E_\datadist\lrs{\margin_\posterior(X,Y)}$, the second moment is given by
$\margin_{\posterior^2} = \E_\datadist\lrs{\lr{\margin_\posterior(X,Y)}^2} = \E_{\hyp_1,\hyp_2\sim\rho^2}\lrs{\E_\datadist\lrs{\hyp_1(X)\hyp_2(X)}}$.
Then the  \cbound\ for the expected loss of the $\posterior$-weighted majority vote classifier reads:
\begin{theorem}[\cbound, \citealp{Germain2015}] 
  For any distribution $\posterior$ over $\Hspace$ and any distribution $\datadist$ on $\Xspace\times \{-1,1\}$, if $\margin_\posterior > 0$, we have
  \begin{equation*}
    \label{eq:c-bound}
    L(\mvoter) \leq 1 - \frac{\margin_{\rho}^2}{\margin_{\rho^2}}.
  \end{equation*}
  \label{thm:c-bound}
\end{theorem}
The theorem follows from  the one-sided Chebyshev inequality applied to the loss of $\mvoter$. 
As the first and second moments are usually not known, Germain et al.~offer several ways to bound them empirically.
They start by showing that
\begin{equation}
  \margin_\posterior = 1-2 \gloss \hspace{10mm} 
  \label{eq:margin1st}
\end{equation}
meaning that the first moment of the margin can be bounded by the use of the \pbbound\ (Theorem~\ref{thm:PBkl}). For the second moment, we have that
\begin{equation}
  \margin_{\posterior^2} = 1-2 \disagree,
  \label{eq:margin2nd}
\end{equation}
where
$\disagree = \frac{1}{2}\lrs{1-\E_\datadist\lrs{\lr{\E_\posterior\lrs{\hyp(X)}}^2}}$
is the \emph{disagreement} between individual classifiers.
Together with the \cbound, the relations above confirm
the observations made by \citet{Breiman2001}: The strength of
$\mvoter$ depends on having strong individual classifiers (low
$\gloss$, i.e., large $\margin_\posterior$) and low correlation
between classifiers (high disagreement $\disagree$, i.e., low $\margin_{\posterior^2}$).

By \eqref{eq:margin1st}, the first moment of the margin can be bounded by the use of the \pbbound\ (Theorem~\ref{thm:PBkl}), while by \eqref{eq:margin2nd} the second moment can be bounded using a lower bound on $\disagree$. 
With $\disagreeemp^\set$ denoting the empirical disagreement computed on $\set$, $\disagree$ can be lower bounded by the smallest $d$ satisfying \citep{Germain2015}
\begin{equation*}
  \kl\lr{\disagreeemp^{\set} \middle\| d} \leq \frac{2\KL\lr{\posterior\middle\|\prior}+\ln \frac{\xi(\tsize)}{\delta}}{\tsize}.
\end{equation*} 
Like in the case of Theorem~\ref{thm:PBkl}, solutions to the above inequality can be computed by a root-finding method. This leads to the following empirical \cbound, which we denote the \cboundempA:

\newcommand{\rr}{b}
\begin{theorem}[\cboundempA, \citealp{Germain2015}] \label{thm:c1-bound}
  For any probability distribution $\prior$ over $\Hspace$ that is independent of $\set$ and any $\delta \in \lr{0,1}$, with probability at least $1-\delta$ over a random draw of a sample $\set$, for all distributions $\posterior$ over $\Hspace$ simultaneously
  \begin{equation*}
    \label{eq:c1-bound}
    \mloss \leq 1 - \frac{\lr{1-2\rr}^2}{(1-2d)}.
  \end{equation*}
  Here $\rr$ is an upper bound on $\gloss$, which can be found by Theorem~\ref{thm:PBkl}, and $d$ is a lower bound on $\disagree$.
\end{theorem}
The \cboundempA\ allows direct bounding of $\mloss$. However, {\citet{Germain2015}} also provide another bound based on Theorem~\ref{thm:c-bound}, which does not require bounding $\gloss$ and $\disagree$ separately.
First, we let
$\jerr_\posterior = \E_{h_1,h_2\sim\posterior^2}\lrs{\E_\datadist\lrs{\indicator{h_1(X)\neq Y}\indicator{h_2(X)\neq Y}}}$
denote the \emph{expected joint error}
and $\jerremp^{\set}$ denote the empirical joint error computed on $\set$. Then
the loss of the associated Gibbs classifier can be written as $\gloss = \frac{1}{2}\lr{2\jerr_\posterior+\disagree}$.
The next bound is then based on bounding $\disagree$ and $\jerr_\posterior$ simultaneously, by bounding the $\KL$-divergence between two \emph{trivalent} random variables. A variable $X$ is trivalent if $\P\lr{X=x_1} = p_1$, $\P\lr{X=x_2} = p_2$ and $\P\lr{X=x_3}=1-p_1-p_2$, and similar to $\kl\lr{\cdot\middle\|\cdot}$, $\kl\lr{p_1,p_2\middle\|\, q_1,q_2}$ denotes the $\KL$-divergence between two trivalent random variables with parameters $(p_1,p_2,1-p_1-p_2)$ and $(q_1,q_2,1-q_1-q_2)$.

Using the above and a generalization of the PAC-Bayes inequality to trivalent random variables,
Germain et al.~derive the following bound, which we refer to as the \cboundempB:
\begin{theorem}[\cboundempB, \citealp{Germain2015}]\label{thm:cboundempB}
  For any probability distribution $\prior$ over $\Hspace$ independent of $\set$ and any $\delta \in \lr{0,1}$, with probability at least $1-\delta$ over a random draw of a sample $\set$, for all distributions $\posterior$ over $\Hspace$ simultaneously 
  \begin{equation*}
    \label{eq:c2-bound}
    \mloss \leq \sup_{d,e}\lr{1 - \frac{\lr{1-\lr{2e+d}}^2}{1-2d}},
  \end{equation*}
  where the supremum is over all $d$ and $e$ satisfying
  \begin{align}
    \kl\lr{\disagreeemp^\set,\jerremp^\set\middle\|\,d,e} \leq \frac{2\KL\lr{\posterior\middle\|\prior}+\ln \frac{\xi(\tsize)+\tsize}{\delta}}{\tsize},
    \hspace{5mm}
    d \leq 2\lr{\sqrt{e}-e},
    \hspace{5mm}
    2e+d <1.
    \label{eq:trivalent-kl}
  \end{align}
\end{theorem}
Again, we need to make adjustments in order to apply the \cboundempA\ and \cboundempB\ in the bagging setting. 
When lower bounding the disagreement and the joint error in Theorem \eqref{thm:cboundempB}, we consider the empirical disagreement $\disagreeemp^{\trainset}$ (and joint error $\jerremp^{\trainset}$) between $\hyp_i$ and $\hyp_j$ estimated on $\oobvalset_i \cap \oobvalset_j$ and choose $\tsize = \min_{i,j}\lr{|\oobvalset_i\cap\oobvalset_j|}$ accordingly.

\subsection{Optimizing the Posterior Distribution.}
Aside from providing guarantees on expected performance, PAC-Bayesian bounds can be used to tune classifiers. The prior distribution $\prior$ must be chosen before observing the data, but we are free to choose the posterior distribution $\posterior$ afterwards, for instance one could choose $\posterior$ such that the empirical loss $\mlossemp{\set}$ is minimized.

Breiman has applied \emph{boosting} \citep{Schapire1999}
to random forests in order to optimize the weighting of the vote, finding that it improved the accuracy in some cases \citep{Breiman2001}. We instead consider optimization of the posterior by minimizing the theoretical bounds \citep{Thiemann2017, Germain2015}. However, none of the bounds provided above can easily be used to directly optimize $\rho$, because they are non-convex in $\rho$ \citep{Thiemann2017}. 
\citet{Thiemann2017} and \citet{Germain2015} came up with two different ways to resolve the convexity issue.

\citeauthor{Thiemann2017} apply a relaxation of Theorem~\ref{thm:PBkl} based on Pinsker's inequality, which leads to the following result that we refer to as the \lambdabound:
\begin{theorem}[\lambdabound, \citealp{Thiemann2017}] 
  For any probability distribution $\pi$ over ${\cal H}$ that is independent of $\set$ and any $\delta \in (0,1)$, with probability at least $1-\delta$ over a random draw of a sample $\set$, for all distributions $\rho$ over ${\cal H}$ and $\lambda \in (0,2)$ simultaneously
  \begin{equation}
    \label{eq:PBlambda}
    \gloss \leq \frac{\glossemp{\set}}{1 - \frac{\lambda}{2}} + \frac{\KL(\rho\|\pi) + \ln \frac{2 \sqrt n}{\delta}}{\lambda\lr{1-\frac{\lambda}{2}}n}.
  \end{equation}
  \label{thm:PBlambda}
\end{theorem}
They show that the \lambdabound\ is convex in $\posterior$ and in $\lambda$ (but not jointly convex). They give an iterative update procedure, which alternates between updating $\lambda$ and $\rho$, and prove that, under certain conditions, the procedure is guaranteed to converge to the global minimum.

\cite{Germain2015} state a version of the \cbound\ that is suited for optimization of $\posterior$.
However, the bound is restricted to \emph{self-complemented hypothesis classes} and posteriors \emph{aligned} on the prior. A hypothesis class is said to be self-complemented, if $\hyp \in \Hspace \Leftrightarrow -\hyp \in \Hspace$, where $-h$ is a hypothesis that always predicts the opposite of $h$ (in binary prediction). A posterior $\posterior$ is said to be aligned on $\prior$ if
$\posterior(\hyp) + \posterior(-\hyp) = \prior(\hyp) + \prior(-\hyp)$.
Thus, the final statement of the bound{, which we denote by \emph{\cboundempC},} becomes:
\begin{theorem}[\cboundempC, \citealp{Germain2015}] 
  For any self-complemented hypothesis set $\Hspace$, any probability distribution $\prior$ over $\Hspace$ independent of $\set$ and any $\delta \in \lr{0,1}$, with probability at least $1-\delta$ over a random draw of a sample $\set$, for all distributions $\posterior$ aligned with $\prior$ simultaneously
  \begin{equation*}
    \mloss \leq 1-\frac{\lr{1-2r}^2}{(1-2d)},
  \end{equation*}
  where
  \begin{equation*}
    r = \min\lr{\frac{1}{2},\glossemp{\set}+\sqrt{\frac{\ln \frac{2\xi(\tsize)}{\delta}}{2\tsize}}},
    \hspace{6mm}
    d = \max\lr{0,\disagreeemp^\set-\sqrt{\frac{\ln \frac{2\xi(\tsize)}{\delta}}{2\tsize}}}.
  \end{equation*}
  \label{thm:Cboundopt}
\end{theorem}
The authors show how to minimize the bound in Theorem~\ref{thm:Cboundopt} over the posterior $\posterior$, by solving a quadratic program. The quadratic program requires a hyperparameter $\mu$, used to {enforce a minimum} value of the first moment of the margin. $\mu$ can be chosen by cross validation \citep{Germain2015}. Furthermore, they note how the restriction to aligned posteriors acts as regularization.

For both the \lambdabound\ and the \cboundempC, we need to make the
same adjustments as for the \pbbound\ and the \cbound{, that is, we choose $\tsize = \min_{i,j}\lr{|\oobvalset_i\cap\oobvalset_j|}$}.
When applying the optimization procedure of the \cboundempC, we also need to make sure that the $\Hspace$ is self-complemented; given a set of hypotheses, this can be done by copying all hypotheses and inverting their predictions.

\begin{table}[ht]
  \caption{Overview of the bounds that we apply in Section~\ref{section:experiments} with references to the corresponding theorems in Section~\ref{section:bounds}. 
  \label{table:bounds-overview}}
  \centering
  \begin{tabular}{llll}
    \hline
    Name & Bound & Theorem & Reference \\
    \hline
    \hline
    \pbbound & $\kl\lr{\glossemp{\set}\middle\|\gloss} \leq \frac{\KL(\rho\|\pi) + \ln \frac{2 \sqrt \tsize}{\delta}}{\tsize}$ & Thm.~\ref{thm:PBkl} & \citealp{Seeger2002} \\
    \hline
    \cboundempA & $\mloss \leq 1 - \frac{\lr{1-2\rr}^2}{(1-2d)}$ & Thm.~\ref{thm:c1-bound} & \citealp{Germain2015} \\
    \hline
    \cboundempB & $\mloss \leq \sup_{d,e}\lr{1 - \frac{\lr{1-\lr{2e+d}}^2}{1-2d}}$ & Thm.~\ref{thm:cboundempB} & \citealp{Germain2015} \\
    \hline
    \shbound & $\kl\lr{\mlossemp{\set}\middle\|\mloss} \leq \frac{\ln \frac{2 \sqrt \tsize}{\delta}}{\tsize}$ & & PBkl with $|{\cal H}|=1$\\
    \hline
    \lambdabound & $\gloss \leq \frac{\glossemp{\set}}{1 - \frac{\lambda}{2}} + \frac{\KL(\rho\|\pi) + \ln \frac{2 \sqrt n}{\delta}}{\lambda\lr{1-\frac{\lambda}{2}}n}$ & Thm.~\ref{thm:PBlambda} & \citealp{Thiemann2017} \\
    \hline
    \cboundempC & $\mloss \leq 1-\frac{\lr{1-2r}^2}{(1-2d)}$ & Thm.~\ref{thm:Cboundopt} & \citealp{Germain2015} \\
\hline
  \end{tabular}
\end{table}

\section{Experiments}
\label{section:experiments}

We have applied the bounds of Section~\ref{section:bounds}, summarized in Table~\ref{table:bounds-overview}, in different random forest settings.
First, we considered the \textbf{standard setting with bagging} and used 
the sets $\oobvalset_1, \oobvalset_2, ...$ for evaluation and computation of the bounds as described in Section~\ref{section:bounds}. The posterior distribution $\posterior$ was taken uniform and not optimized.
Then we considered a \textbf{setting with a separate validation set} $\valset$.
The majority vote bounds suffer from the low number of training patterns available for evaluating the correlation between classifiers. 
Therefore, we also evaluated the quality of the bounds when a separate validation set $\valset$ was set aside before training. $\valset$ was then used in addition to $\oobvalset_1,\oobvalset_2,...$, when evaluating and computing the bounds. Again, the posterior distribution $\posterior = \uniform$ was not optimized.
Finally, we looked at \textbf{random forests with optimized posteriors}.
We used bagging and the bounds in Theorems~\ref{thm:PBlambda} and \ref{thm:Cboundopt} to optimize the posterior distribution $\posterior$.

{For all settings, the accuracy of the final majority vote classifier $\mvoter$ is also of interest. Hence, a separate test set $\testset$ is left out in each setting. This set is used only for evaluating the final classifier by $\mlossemp{\testset}$.}
{We are mainly concerned with the tightness of the bounds when
  individual voters are strong. {Therefore}, all features are
  considered in each split during the training of the random forest
  (using Gini impurity as splitting {criterion}), and trees are
  trained until all leaves are pure {(see, e.g., \citealp{gieseke:18},
    for arguments why this can be beneficial)}.}

  {To study  how the bounds depend on the strengths of the individual
    classifiers, we varied the maximum tree depth and the number of
    features considered in each split  in the first two settings.
    This allows {us} to  investigate the evolution of the bounds
 as the strength of individual classifiers increases by going from
 decision stumps to full-grown trees. }
{We either set the number of random features considered for splitting
to the maximum number or, to further weaken the classifiers, to a
single random feature. We restricted these experiments to two data sets.}

Experiments were run on several binary UCI data sets {(see left part of Table~\ref{tab:results-bagtest}).}
For each data set, all patterns with one or more missing
features were removed.
Since the \cbound\ is only analysed for binary classification, we restrict ourselves to binary tasks.
The number of trees $\hcount$ for any data set of size $\datasetsize$ was chosen as the largest value from $\{100,200,500,1000\}$ that was smaller than {$\datasetsize/4$}.

  For each setting, $\datasetsize/2$ patterns were randomly sampled
  for $\testset$. In the first and third settings, all remaining
  patterns were used for training. In the second setting, a further
  $\datasetsize/4$ patterns were sampled for $\valset$, with the
  remaining patterns used for training, see
    Figure~\ref{fig:data-overview} for an illustration.

When evaluating the bounds, we chose $\prior = \uniform$ and $\delta = 0.05$. 

\begin{landscape}
  \begin{table}
    \caption{
      The \pbbound, \cboundempA, \cboundempB\ and \shbound\ computed
      for the binary UCI data sets in the bagging and validation set
      settings. In both settings, the majority vote loss on $\testset$
      is given as an estimate of the accuracy of the trained
      classifier denoted as \emph{test score}.
      The best bound within an experiment is marked \textbf{bold}, while \textit{italics} is used to indicate trivial bounds ($\geq 0.5$).
    }
    \label{tab:results-bagtest}
    \vskip 0.1in
    \begin{center}
      \begin{small}
\begin{sc}
\begin{tabular}{lcc|cccc|ccccc}
  \toprule
  \multicolumn{3}{l}{} & \multicolumn{4}{|l}{Bagging} & \multicolumn{4}{|l}{Validation set} \\
	Data set & $n$ & $d$ & Test score & \PB & $\C1$ & $\C2$ & Test score & \PB & $\C1$ & $\C2$ & $\sinhyp$ \\
  \midrule
  Adult & 45222 & 14 & 0.152 & \textbf{0.428} & \textit{0.525} & \textit{0.520} & 0.154 & 0.426 & 0.500 & 0.479 & \textbf{0.169}\\
Credit-A & 653 & 15 & 0.144 & \textbf{\textit{0.603}} & \textit{0.853} & \textit{0.990} & 0.135 & \textit{0.563} & \textit{0.790} & \textit{0.788} & \textbf{0.294}\\
Haberman & 306 & 3 & 0.333 & $>$\textit{1} & $>$\textit{1} & $>$\textit{1} & 0.333 & $>$\textit{1} & $>$\textit{1} & $>$\textit{1} & \textbf{\textit{0.577}}\\
Heart & 297 & 13 & 0.228 & \textbf{\textit{0.960}} & $>$\textit{1} & $>$\textit{1} & 0.282 & \textit{0.822} & \textit{0.973} & \textit{0.986} & \textbf{0.412}\\
ILPD & 579 & 10 & 0.307 & $>$\textit{1} & $>$\textit{1} & $>$\textit{1} & 0.307 & \textit{0.955} & \textit{0.999} & $>$\textit{1} & \textbf{0.441}\\
Ionosphere & 351 & 34 & 0.108 & \textbf{\textit{0.691}} & \textit{0.920} & $>$\textit{1} & 0.125 & \textit{0.649} & \textit{0.878} & \textit{0.890} & \textbf{0.299}\\
Letter:AB & 20000 & 16 & 0.010 & \textbf{0.124} & 0.244 & 0.408 & 0.015 & 0.140 & 0.242 & 0.186 & \textbf{0.035}\\
Letter:DO & 20000 & 16 & 0.045 & \textbf{0.216} & 0.391 & \textit{0.530} & 0.067 & 0.227 & 0.362 & 0.294 & \textbf{0.072}\\
Letter:OQ & 20000 & 16 & 0.051 & \textbf{0.288} & 0.491 & \textit{0.605} & 0.059 & 0.323 & 0.474 & 0.404 & \textbf{0.119}\\
Mushroom & 8124 & 22 & 0.000 & \textbf{0.011} & 0.025 & 0.078 & 0.001 & 0.013 & 0.028 & 0.037 & \textbf{0.008}\\
Sonar & 208 & 60 & 0.221 & $>$\textit{1} & $>$\textit{1} & $>$\textit{1} & 0.250 & $>$\textit{1} & $>$\textit{1} & $>$\textit{1} & \textbf{\textit{0.510}}\\
Tic-Tac-Toe & 958 & 9 & 0.088 & \textbf{\textit{0.627}} & \textit{0.847} & \textit{0.934} & 0.142 & \textit{0.749} & \textit{0.892} & \textit{0.805} & \textbf{0.221}\\
USvotes & 232 & 16 & 0.034 & \textbf{\textit{0.526}} & \textit{0.810} & $>$\textit{1} & 0.052 & 0.497 & \textit{0.764} & \textit{0.750} & \textbf{0.228}\\
WDBC & 569 & 30 & 0.053 & \textbf{0.430} & \textit{0.696} & \textit{0.945} & 0.063 & 0.327 & \textit{0.543} & 0.489 & \textbf{0.102}\\

  \bottomrule
\end{tabular}
\end{sc}
\end{small}


    \end{center}
    \vskip 0.1in
  \end{table}
  \begin{figure}
    \begin{center}
      \begin{tikzpicture}
  \draw [anchor=south west] (0, 0.9) node {\textsc{Bagging}};
  \draw [draw=black,thick] (0, 0)  rectangle ++(2.5,0.8) node[pos=.5] {$\trainset$};
  \draw [draw=black,thick] (2.5, 0)  rectangle ++(2.5,0.8) node[pos=.5] {$\testset$};
  \draw [decorate,decoration={brace,amplitude=8pt}]
  (2.5,-0.1) -- (0,-0.1)
  node [anchor=north,text width=2.5cm,black,midway,align=center,yshift=-0.3cm]
  {\footnotesize Training + $\PB,\C1,\C2$};
  \draw [decorate,decoration={brace,amplitude=8pt}]
  (5,-0.1) -- (2.5,-0.1)
  node [anchor=north,text width=2.5cm,black,midway,align=center,yshift=-0.3cm]
  {\footnotesize Testing};
  %
  \draw [anchor=south west] (6, 0.9) node {\textsc{Validation set}};
  \draw [draw=black,thick] (6,  0)  rectangle ++(1.25,0.8) node[pos=.5] {$\trainset$};
  \draw [draw=black,thick] (7.25,0)  rectangle ++(1.25,0.8) node[pos=.5] {$\valset$};
  \draw [draw=black,thick] (8.5, 0)  rectangle ++(2.5,0.8) node[pos=.5] {$\testset$};
  \draw [decorate,decoration={brace,amplitude=8pt,aspect=0.75}]
  (8.5,-0.1) -- (6,-0.1)
  node [anchor=north,text width=1.5cm,black,midway,align=center,xshift=-0.75cm,yshift=-0.3cm]
  {\footnotesize Training + $\PB,\C1,\C2$};
  \draw [decorate,decoration={brace,amplitude=8pt,raise=6pt}]
  (8.5,-0.1) -- (7.25,-0.1)
  node [anchor=north,text width=1.5cm,black,midway,align=center,yshift=-0.5cm]
  {\footnotesize $\sinhyp$};
  \draw [decorate,decoration={brace,amplitude=8pt}]
  (11,-0.1) -- (8.5,-0.1)
  node [anchor=north,text width=3cm,black,midway,align=center,yshift=-0.3cm]
  {\footnotesize Testing};
  %
  \draw [anchor=south west] (12, 0.9) node {\textsc{Optimization}};
  \draw [draw=black,thick] (12, 0)  rectangle ++(2.5,0.8) node[pos=.5] {$\trainset$};
  \draw [draw=black,thick] (14.5, 0)  rectangle ++(2.5,0.8) node[pos=.5] {$\testset$};
  \draw [decorate,decoration={brace,amplitude=8pt}]
  (14.5,-0.1) -- (12,-0.1)
  node [anchor=north,text width=2.5cm,black,midway,align=center,yshift=-0.3cm]
  {\footnotesize Training + Optimization};
  \draw [decorate,decoration={brace,amplitude=8pt}]
  (17,-0.1) -- (14.5,-0.1)
  node [anchor=north,text width=2.5cm,black,midway,align=center,yshift=-0.3cm]
  {\footnotesize Testing};
\end{tikzpicture}
    \end{center}
    \caption{Overview of data split for each of the three settings. Note, that the bounds computed on $\trainset$, as well as the optimization in the third setting, are only based on the hold-out sets, $\oobvalset_i$.}
    \label{fig:data-overview}
  \end{figure}
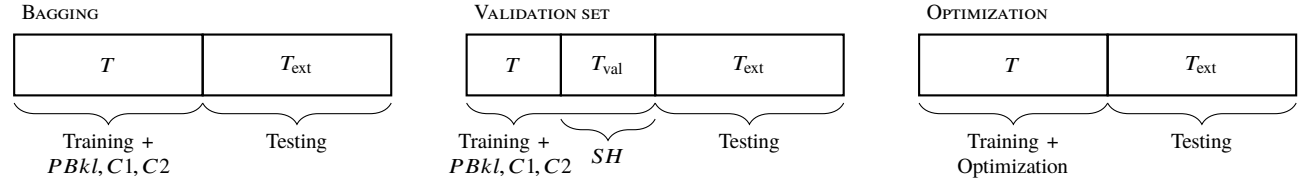  
\end{landscape}

\subsection{Random forest with bagging.}
We started with the original random forest setting, where an individual tree $\hyp_i$ is trained on a bootstrap sample $\trainset_i$ of size $|\trainset|$, drawn with replacement from the training set $\trainset$
consisting of half the data with the other half $\testset$ used for evaluating the final classifier.
As mentioned, the posterior distribution $\posterior$ was chosen uniform.
In this experiment, $\trainset$ comprised all available data. The empirical Gibbs loss was  evaluated using  $\oobvalset_i = \trainset \setminus \trainset_i$ and the empirical disagreement and joint error between two trees $\hyp_i$ and $\hyp_j$ using $\oobvalset_i \cap \oobvalset_j$.

We considered  the \pbbound\ and the two empirical \cbound s, \cboundempA\ and \cboundempB, with sample sizes $n$ calculated as described in 
Section~\ref{section:bounds}.
  Furthermore, the trained classifier $\mvoter$ was evaluated on $\testset$.

Table~\ref{tab:results-bagtest} (middle) lists the results.
The test score $\mlossemp{\testset}$ provides an estimate of the accuracy of the classifier.
The \pbbound\ always gave the tightest  bounds. For {6} out of the 14 data sets the bound was below 0.5.
The better performance of the  \pbbound\ is explained by the high accuracy of the individual trees.
As mentioned by \cite{Germain2015} and discussed in Section~\ref{section:bounds}, the \cbound\ degrades when the individual classifiers are strong. Thus, the \pbbound\ including the factor of 2 coming from \eqref{eq:factor2} is tighter. Furthermore, for the \cbound s the bagging setting is particularly difficult, because  there is only  a small amount of data available for estimating correlations. This is especially true for the \cboundempB, since it relies only on the intersection between the two samples $\oobvalset_i$ and $\oobvalset_j$, which may be small. 

In the bagging setting we get the bounds ``for free'' in the sense that  all evaluations are based on the $\oobvalset_i$ sets, which are by-products of the training, and we do not have to set aside a separate validation set. Thus, more data is available for selecting the hypothesis.

\begin{figure}[ht]
  \centering
  \begin{tikzpicture}
  \begin{groupplot}[
      group style={group size= 2 by 2, vertical sep=1.3cm},
      height=5.6cm,width=6.4cm,
    ]
    \nextgroupplot[ComplexFig,title={\dataset{Letter:AB}, best feature},legend to name=plotleg4,legend style={legend columns=5,/tikz/every even column/.append style={column sep=0.4cm}}]
    \addplot[ComplexTest] table [x=gibbs, y=test, col sep=semicolon]{figs/complex/complex-noval-letterAB-full.csv};
    \addlegendentry{$\mlossemp{\testset}$};
    \addplot[ComplexDisagree] table [x=gibbs, y=disagree, col sep=semicolon]{figs/complex/complex-noval-letterAB-full.csv};
    \addlegendentry{$\disagreeemp$};
    \addplot[ComplexPB] table [x=gibbs, y=p0, col sep=semicolon]{figs/complex/complex-noval-letterAB-full.csv};
    \addlegendentry{\PB};
    \addplot[ComplexC1] table [x=gibbs, y=p1, col sep=semicolon]{figs/complex/complex-noval-letterAB-full.csv};
    \addlegendentry{\C1};
    \addplot[ComplexC2] table [x=gibbs, y=p2, col sep=semicolon]{figs/complex/complex-noval-letterAB-full.csv};
    \addlegendentry{\C2};
    \nextgroupplot[ComplexFig,title={\dataset{Letter:AB}, random feature}]
    \addplot[ComplexTest] table [x=gibbs, y=test, col sep=semicolon]{figs/complex/complex-noval-letterAB.csv};
    \addplot[ComplexDisagree] table [x=gibbs, y=disagree, col sep=semicolon]{figs/complex/complex-noval-letterAB.csv};
    \addplot[ComplexPB] table [x=gibbs, y=p0, col sep=semicolon]{figs/complex/complex-noval-letterAB.csv};
    \addplot[ComplexC1] table [x=gibbs, y=p1, col sep=semicolon]{figs/complex/complex-noval-letterAB.csv};
    \addplot[ComplexC2] table [x=gibbs, y=p2, col sep=semicolon]{figs/complex/complex-noval-letterAB.csv};
    \nextgroupplot[ComplexFig,title={\dataset{Mushroom}, best feature}]
    \addplot[ComplexTest] table [x=gibbs, y=test, col sep=semicolon]{figs/complex/complex-noval-mushroom-full.csv};
    \addplot[ComplexDisagree] table [x=gibbs, y=disagree, col sep=semicolon]{figs/complex/complex-noval-mushroom-full.csv};
    \addplot[ComplexPB] table [x=gibbs, y=p0, col sep=semicolon]{figs/complex/complex-noval-mushroom-full.csv};
    \addplot[ComplexC1] table [x=gibbs, y=p1, col sep=semicolon]{figs/complex/complex-noval-mushroom-full.csv};
    \addplot[ComplexC2] table [x=gibbs, y=p2, col sep=semicolon]{figs/complex/complex-noval-mushroom-full.csv};
    \nextgroupplot[ComplexFig,title={\dataset{Mushroom}, random feature}]
    \addplot[ComplexTest] table [x=gibbs, y=test, col sep=semicolon]{figs/complex/complex-noval-mushroom.csv};
    \addplot[ComplexDisagree] table [x=gibbs, y=disagree, col sep=semicolon]{figs/complex/complex-noval-mushroom.csv};
    \addplot[ComplexPB] table [x=gibbs, y=p0, col sep=semicolon]{figs/complex/complex-noval-mushroom.csv};
    \addplot[ComplexC1] table [x=gibbs, y=p1, col sep=semicolon]{figs/complex/complex-noval-mushroom.csv};
    \addplot[ComplexC2] table [x=gibbs, y=p2, col sep=semicolon]{figs/complex/complex-noval-mushroom.csv};
  \end{groupplot}
  \node[anchor=north] (legend) at ($(group c1r2.south east)!0.5!(group c2r2.south west)-(0,0.5cm)$) {\ref{plotleg4}};
\end{tikzpicture}
  \caption{Evolution of bounds depending on voter strength (measured
    by Gibbs risk) on data sets \dataset{Letter:AB} (top) and
    \dataset{Mushroom} (bottom) in the bagging setting, using the best
    feature for splits (left) or a random feature (right). {In
      addition, the {empirical} disagreement $\disagreeemp$ between the trees is plotted.}}
  \label{fig:complex-setting1}
\end{figure}
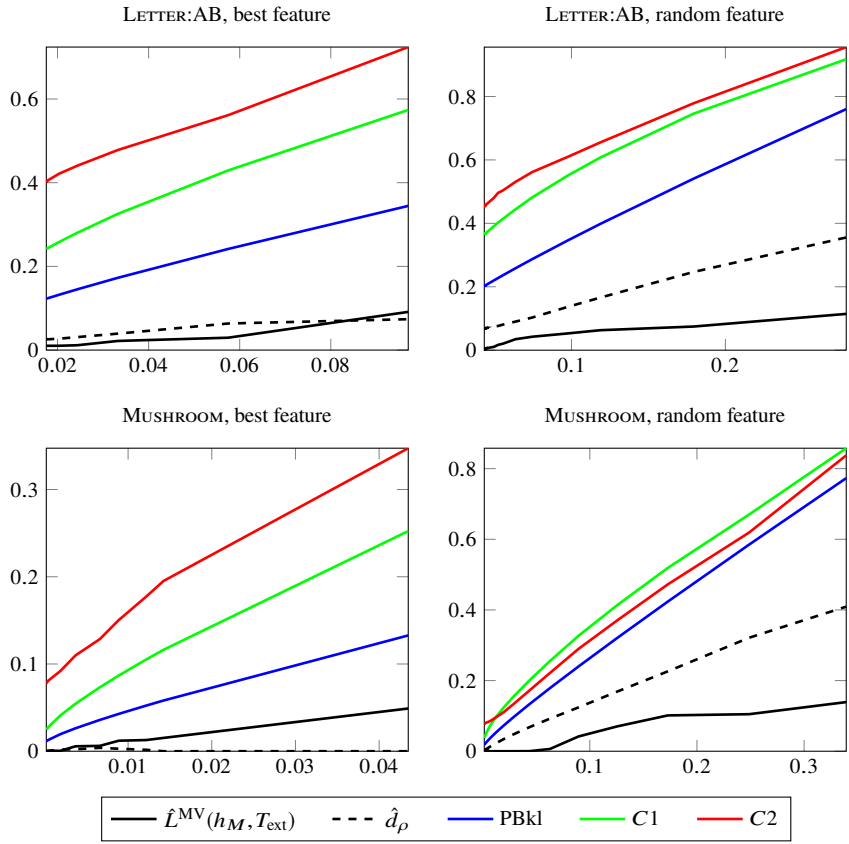

{Figure~\ref{fig:complex-setting1} shows the evolution of the
  bounds as the strength of the individual voters varies for the data
  sets \dataset{Letter:AB} and \dataset{Mushroom}. Voter strength was
  controlled by {increasing} the maximum
  allowed tree depth until only pure trees were obtained, and by
  feature selection during splits, that is, using either the best feature (stronger voters) or a random feature (weaker voters).}

{From the figure, we see that the \pbbound\ is tighter than both
  the \cboundempA\ and the \cboundempB, even though the \C-bounds are
  expected to perform better, when individual classifiers are
  weak. However, this theoretical advandtage is {outweighed} by the low
  amount of data available for bounding the
  disagreement/joint error, that is, $\tsize = \min_{i,j}\lr{|\oobvalset_i\cap\oobvalset_j|}$ is very small, leading to loose bounds.}

\subsection{Random forest with a separate validation set.}
As a reference,  we considered the scenario where a separate
validation set $\valset$ was set aside before the random forest was
trained, which allows for a better estimate of the correlations in the
\cbound s. Recall that a separate test set $\testset$ was set
aside for evaluating the classifier beforehand. Now the remaining
half of the
data set was split into two equal sized parts, $\trainset$ and
$\valset$. The random forest was then trained on $\trainset$ as before
using bagging, but the empirical Gibbs loss and disagreement were now
measured on the sets $\oobvalset_1,\oobvalset_2,...$ combined with
$\valset$. {We also considered the setting in which only $\valset$ was used for computing the bounds. This, as expected, led to slightly worse bounds. The results can be found in {Table~\ref{tab:results-bagtest-no-oob} in} the appendix.} 
As in the previous setting, we had to take care when applying the bounds.
Again, the sample sizes $n$ for the theorems were calculated as described in Section~\ref{section:bounds}, but now with extra $\datasetsize/4$ points available.
{As before}, we applied the \pbbound, the \cboundempA\ and the
\cboundempB, but now with the addition of the
  \shbound. Having the separate validation set allowed us to
  apply this single hypothesis bound, which is based only on $\valset$.

Table~\ref{tab:results-bagtest} (right) lists the results.
{Again, the loss of $\mvoter$ on $\testset$ is given as an estimate of the {accuracy} of the classifier.}
As before, we see that the \pbbound\ was tighter than the \cbound s in almost all cases, and again the explanation lies in the strength of the individual classifiers. We also see that the \cboundempB\ was tighter than \cboundempA. 
This is in accordance with the observation by \cite{Germain2015} that the \cboundempB\ is often tighter
when there is an equal amount of data available for estimating the empirical Gibbs loss and the empirical correlation between any two classifiers.
{However, we see in all cases that the single hypothesis bound gives the best guarantees. This indicates that the \pbbound\ does indeed suffer from not taking correlations into account, even if it outperforms the \cbound s.}

Comparing the results to the bounds obtained in the previous experiment, we see that, with the exception of the \shbound, the bounds overall were very similar, some bounds better, some worse.  This can be explained by the trade-off between using data for training the classifier and using data for evaluating the classifier as part of computing the bounds.  In the previous experiment, more data was used to train the random forest, which typically gives a better classifier (as also indicated by the performance on the test set $\testset$), resulting in a lower empirical Gibbs loss.  Still, in this experiment the bound can be tighter because more data is used to evaluate the classifiers.  {This is demonstrated in Figure~\ref{fig:comparison} in
  Appendix~\ref{app:comparison} which shows a comparison of the two
  settings on two exemplary data sets, \dataset{Letter:DO} and
  \dataset{Adult}. The figure illustrates the difference in tightness
  of the bounds.}

The \shbound\ provides the best guarantees for all data sets across both experiments, indicating that the other bounds are still too loose. The \shbound\ does not come for free though, as data must be set aside, whereas the bounds computed in the bagging setting often provide useful guarantees and a better classifier.

\begin{figure}[ht]
  \centering
  \begin{tikzpicture}
  \begin{groupplot}[
      group style={group size= 2 by 2, vertical sep=1.3cm},
      height=5.6cm,width=6.4cm,
    ]
    \nextgroupplot[ComplexFig,title={\dataset{Letter:AB}, best feature},legend to name=plotleg3,legend style={legend columns=6,/tikz/every even column/.append style={column sep=0.4cm}}]
    \addplot[ComplexTest] table [x=gibbs, y=test, col sep=semicolon]{figs/complex/complex-val-letterAB-full.csv};
    \addlegendentry{$\mlossemp{\testset}$};
    \addplot[ComplexDisagree] table [x=gibbs, y=disagree, col sep=semicolon]{figs/complex/complex-val-letterAB-full.csv};
    \addlegendentry{$\disagreeemp$};
    \addplot[ComplexPB] table [x=gibbs, y=p0, col sep=semicolon]{figs/complex/complex-val-letterAB-full.csv};
    \addlegendentry{\PB};
    \addplot[ComplexC1] table [x=gibbs, y=p1, col sep=semicolon]{figs/complex/complex-val-letterAB-full.csv};
    \addlegendentry{\C1};
    \addplot[ComplexC2] table [x=gibbs, y=p2, col sep=semicolon]{figs/complex/complex-val-letterAB-full.csv};
    \addlegendentry{\C2};
    \addplot[ComplexSH] table [x=gibbs, y=psing, col sep=semicolon]{figs/complex/complex-val-letterAB-full.csv};
    \addlegendentry{\sh};
    \nextgroupplot[ComplexFig,title={\dataset{Letter:AB}, random feature}]
    \addplot[ComplexTest] table [x=gibbs, y=test, col sep=semicolon]{figs/complex/complex-val-letterAB.csv};
    \addplot[ComplexDisagree] table [x=gibbs, y=disagree, col sep=semicolon]{figs/complex/complex-val-letterAB.csv};
    \addplot[ComplexPB] table [x=gibbs, y=p0, col sep=semicolon]{figs/complex/complex-val-letterAB.csv};
    \addplot[ComplexC1] table [x=gibbs, y=p1, col sep=semicolon]{figs/complex/complex-val-letterAB.csv};
    \addplot[ComplexC2] table [x=gibbs, y=p2, col sep=semicolon]{figs/complex/complex-val-letterAB.csv};
    \addplot[ComplexSH] table [x=gibbs, y=psing, col sep=semicolon]{figs/complex/complex-val-letterAB.csv};
    \nextgroupplot[ComplexFig,title={\dataset{Mushroom}, best feature}]
    \addplot[ComplexTest] table [x=gibbs, y=test, col sep=semicolon]{figs/complex/complex-val-mushroom-full.csv};
    \addplot[ComplexDisagree] table [x=gibbs, y=disagree, col sep=semicolon]{figs/complex/complex-val-mushroom-full.csv};
    \addplot[ComplexPB] table [x=gibbs, y=p0, col sep=semicolon]{figs/complex/complex-val-mushroom-full.csv};
    \addplot[ComplexC1] table [x=gibbs, y=p1, col sep=semicolon]{figs/complex/complex-val-mushroom-full.csv};
    \addplot[ComplexC2] table [x=gibbs, y=p2, col sep=semicolon]{figs/complex/complex-val-mushroom-full.csv};
    \addplot[ComplexSH] table [x=gibbs, y=psing, col sep=semicolon]{figs/complex/complex-val-mushroom-full.csv};
    \nextgroupplot[ComplexFig,title={\dataset{Mushroom}, random feature}]
    \addplot[ComplexTest] table [x=gibbs, y=test, col sep=semicolon]{figs/complex/complex-val-mushroom.csv};
    \addplot[ComplexDisagree] table [x=gibbs, y=disagree, col sep=semicolon]{figs/complex/complex-val-mushroom.csv};
    \addplot[ComplexPB] table [x=gibbs, y=p0, col sep=semicolon]{figs/complex/complex-val-mushroom.csv};
    \addplot[ComplexC1] table [x=gibbs, y=p1, col sep=semicolon]{figs/complex/complex-val-mushroom.csv};
    \addplot[ComplexC2] table [x=gibbs, y=p2, col sep=semicolon]{figs/complex/complex-val-mushroom.csv};
    \addplot[ComplexSH] table [x=gibbs, y=psing, col sep=semicolon]{figs/complex/complex-val-mushroom.csv};
  \end{groupplot}
  \node[anchor=north] (legend) at ($(group c1r2.south east)!0.5!(group c2r2.south west)-(0,0.5cm)$) {\ref{plotleg3}};
\end{tikzpicture}
  \caption{Evolution of bounds depending on  voter strength (measured
    by Gibbs risk) on data sets \dataset{Letter:AB} (top) and
    \dataset{Mushroom} (bottom) in the setting with a separate
    validation set, using the best feature for splits (left) or a
    random feature (right). In
      addition, the {empirical} disagreement $\disagreeemp$ between the trees is plotted.}
  \label{fig:complex-setting2}
\end{figure}
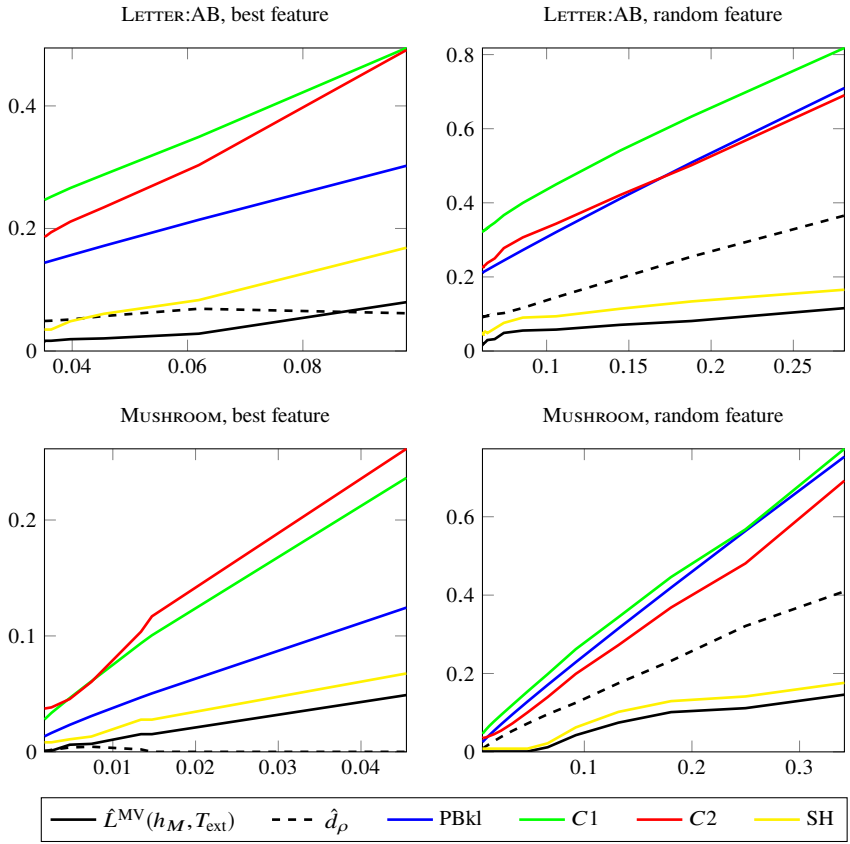

{The dependence of the bounds on  the strengths
  of the individual voters is shown in
  Figure~\ref{fig:complex-setting2} for the data sets
  \dataset{Letter:AB} and \dataset{Mushroom}.} {As in the previous setting, maximum tree depth and feature selection at splits (using the best or a random feature) were used to control voter strength.}

{{Even when individual voters got
   weaker, the \shbound\ remained} tighter. As expected, the \cboundempB\
  now outperformed the \pbbound\ when individual voters are weak and
  disagreement is high.
  {However, to  observe this  effect it was necessary to consider a
    single random feature for splitting.}
 Considering the best feature with shallow trees also results in weak voters, but because of {lower} disagreement (due to trees being very similar), the \C-bounds are still loose.}
 
{  Comparing to the bagging setting, we see the
  impact of having extra data from the left out validation set,
  $\valset$, when evaluating the bounds. {Still}, the \pbbound\ 
  remained tighter {for strong voters}, that is, when the  Gibbs risk is close to zero.}

\subsection{Random Forest with Optimized Posterior.}
Finally, we optimized the posteriors  based on the \lambdabound\ (Theorem~\ref{thm:PBlambda}) and the \cboundempC\ (Theorem~\ref{thm:Cboundopt}). The former was updated by iterative application of the update rules given by \cite{Thiemann2017}. For the latter, we made sure that the hypothesis set is self-complemented \citep{Germain2015} by adding a copy of all trained trees with their predictions reversed. The quadratic program was then solved using the solver CVXOPT \citep{Dahvan2007}.

\begin{table}
  \caption{Loss on $\testset$ obtained when $\posterior$ is chosen by minimizations of the \lambdabound\ ($\posterior=\posterior_\lambda$) and the {\cboundempC}\ ($\posterior=\posterior_{\C}$), compared to loss obtained with $\posterior=\uniform$.
  }
  \label{tab:results-opt}
  \begin{center}
    \begin{small}
\begin{sc}
\begin{tabular}{lccc}
\toprule
Data set & $\uniform$ & $\posterior_\lambda$ & $\posterior_{\C}$ \\
\midrule
Adult & $0.152$ & $0.170$ & $0.152$\\
Mushroom & $0.000$ & $0.000$ & $0.000$\\
Letter:AB & $0.010$ & $0.010$ & $0.010$\\
Letter:DO & $0.044$ & $0.051$ & $0.044$\\
Letter:OQ & $0.051$ & $0.061$ & $0.051$\\
Tic-Tac-Toe & $0.079$ & $0.069$ & $0.086$\\
Credit-A & $0.144$ & $0.153$ & $0.144$\\

\bottomrule
\end{tabular}
\end{sc}
\end{small}

  \end{center}
\end{table}
For each experiment, we split the data set into a  training set
$\trainset$ and an external test set $\testset$ \emph{not used in the
  model building process}\ci,{ see Figure~\ref{fig:data-overview}}.
We only considered larger benchmark data sets, because $\trainset$ and $\testset$ needed to be of sufficient size.
The random forest was then trained using bagging, and the posteriors were then optimized using the individual sets $\oobvalset_1,\oobvalset_2,...$. We selected the hyperparameter $\mu$ for the quadratic program of Theorem~\ref{thm:Cboundopt} that minimized the OOB estimate. 
Once the optimal $\posterior$ was found, the random forest with optimized weights was evaluated on $\testset$. A random forest with uniform posterior was trained and evaluated in the same setting as a baseline.

Table~\ref{tab:results-opt} lists the loss on $\testset$ for the seven largest data sets when optimizing $\posterior$
by minimzation of the \lambdabound\ and \cboundempC.
$\posterior_\lambda$ and $\posterior_{\C}$ denotes the optimal posteriors found using the optimization with the \lambdabound\ and the \cboundempC\ respectively. Note that for $\posterior_{\C}$, the hypothesis set is modified such that it is self-complemented.

For the optimization using the \lambdabound, we see that, except on
the \dataset{Tic-Tac-Toe} data set, the test loss for the optimized
posterior was equal or slightly higher.
The reason is that, because the \lambdabound\ does not consider
interactions between ensemble members, it tends to put most weight
on only a few trees. Thus, the effect of cancellation of errors
vanishes.

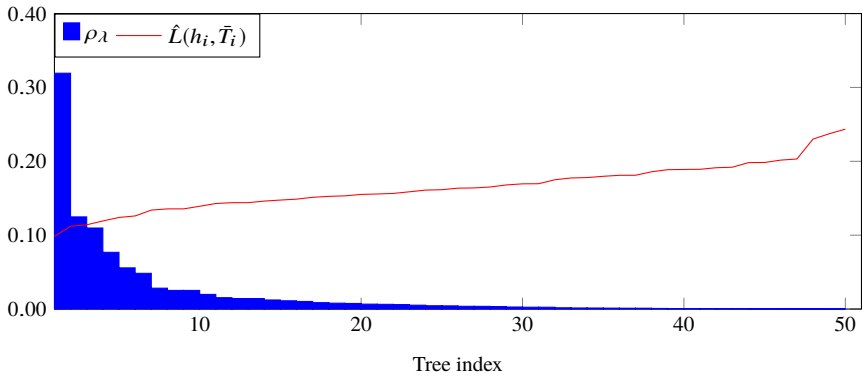
\begin{figure}
  \begin{center}
    \pgfplotstableread[row sep=\\,col sep=&]{
h & risk & rho \\
1 & 0.0983606557377 & 0.319266823759 \\
2 & 0.11214953271 & 0.124744383843 \\
3 & 0.114035087719 & 0.109701112312 \\
4 & 0.119266055046 & 0.0768035359567 \\
5 & 0.123966942149 & 0.0557495136052 \\
6 & 0.126050420168 & 0.0483696392531 \\
7 & 0.133928571429 & 0.0282741682141 \\
8 & 0.135593220339 & 0.0252416882707 \\
9 & 0.135593220339 & 0.0252416882707 \\
10 & 0.139130434783 & 0.0198345093159 \\
11 & 0.142857142857 & 0.0153856468022 \\
12 & 0.144 & 0.0142327452582 \\
13 & 0.144 & 0.0142327452582 \\
14 & 0.146153846154 & 0.0122895968633 \\
15 & 0.147540983607 & 0.0111809878609 \\
16 & 0.148760330579 & 0.0102893773468 \\
17 & 0.151260504202 & 0.00867735196618 \\
18 & 0.152542372881 & 0.00795143605965 \\
19 & 0.153225806452 & 0.0075895640061 \\
20 & 0.155172413793 & 0.00664660373579 \\
21 & 0.155737704918 & 0.00639540285813 \\
22 & 0.15652173913 & 0.00606263630266 \\
23 & 0.15873015873 & 0.00521549067745 \\
24 & 0.161016949153 & 0.00446281780383 \\
25 & 0.161538461538 & 0.0043069821311 \\
26 & 0.163636363636 & 0.00373317115686 \\
27 & 0.16393442623 & 0.00365810053874 \\
28 & 0.165217391304 & 0.0033518265991 \\
29 & 0.168 & 0.00277280897971 \\
30 & 0.169491525424 & 0.00250479820259 \\
31 & 0.169642857143 & 0.00247909700845 \\
32 & 0.175 & 0.00172079475584 \\
33 & 0.177419354839 & 0.00145921512792 \\
34 & 0.177966101695 & 0.00140584140144 \\
35 & 0.1796875 & 0.00125021637352 \\
36 & 0.181034482759 & 0.00114055504018 \\
37 & 0.181102362205 & 0.00113529076439 \\
38 & 0.185840707965 & 0.000821974311593 \\
39 & 0.188679245283 & 0.000677393834634 \\
40 & 0.188976377953 & 0.000663814140912 \\
41 & 0.189189189189 & 0.000654255774538 \\
42 & 0.191304347826 & 0.000566423972054 \\
43 & 0.192 & 0.000540195830001 \\
44 & 0.198198198198 & 0.000354072859091 \\
45 & 0.198412698413 & 0.000348934339083 \\
46 & 0.201612903226 & 0.000280557458617 \\
47 & 0.203125 & 0.000253084603143 \\
48 & 0.230088495575 & 4.0288527365e-05 \\
49 & 0.237288135593 & 2.46650176691e-05 \\
50 & 0.24347826087 & 1.61756517899e-05 \\
}\optA
\begin{tikzpicture}
  \begin{axis}[
      width=\linewidth,
      height=55mm,
      y tick label style={
        /pgf/number format/.cd,
        fixed,
        fixed zerofill,
        precision=2,
        /tikz/.cd
      },
      legend style={at={(0,1)},
        anchor=north west,legend columns=-1},
      xtick={10,20,30,40,50},
      nodes near coords align={vertical},
      ymin=0,ymax=0.4,
      xmin=1,xmax=51,
      xlabel={\small{Tree index}},
      ]
    \addplot[ybar interval, ybar legend,blue,fill=blue] table[x=h,y=rho]{\optA};
    \addplot[red] table[x=h,y=risk]{\optA};
    \legend{$\posterior_\lambda$, {$\hat{L}(h_i,\oobvalset_i)$}}
  \end{axis}
\end{tikzpicture}
  \end{center}
  \caption{Optimized distribution $\posterior_\lambda$ and for each tree $i$ the
    error on the subset $\oobvalset_i$  of the training data not used for
    building the tree.
    Shown are the results for the \dataset{Credit-A} data set using 50 trees.}
  \label{fig:rho-a}
\end{figure}

Figure~\ref{fig:rho-a} demonstrates that indeed  most of the probability mass was centered on the few trees.
However, recomputing the {\pbbound, \cboundempA\ and \cboundempB\ using posterior} $\posterior_\lambda$, we observed that the \pbbound\ (and actually also the \cbound s) became tighter, indicating that the bounds are still quite loose.

Optimizing using the  \cbound\ in Theorem~\ref{thm:Cboundopt} does not
suffer from the probability mass being concentrated on very few tress,
because of the restriction to posteriors aligned on the prior (which
is uniform in our case) and the fact that the individual trees were
rather strong.  The probability mass can only be moved between a tree and its complement.  If $\hyp_i$ has a small loss, $\posterior_\C(\hyp_i)$ is close to $1/\hcount$, since $-\hyp_i$ is very weak.
\begin{figure}
  \begin{center}
    \pgfplotstableread[row sep=\\,col sep=&]{
h & risk & rho \\
1 & 0.0983606557377 & 0.0199999947144 \\
2 & 0.11214953271 & 0.0199999913055 \\
3 & 0.114035087719 & 0.0199999907827 \\
4 & 0.119266055046 & 0.0199999892675 \\
5 & 0.123966942149 & 0.0199999878305 \\
6 & 0.126050420168 & 0.0199999871727 \\
7 & 0.133928571429 & 0.0199999845836 \\
8 & 0.135593220339 & 0.0199999840185 \\
9 & 0.135593220339 & 0.0199999840185 \\
10 & 0.139130434783 & 0.0199999828003 \\
11 & 0.142857142857 & 0.0199999814952 \\
12 & 0.144 & 0.0199999810912 \\
13 & 0.144 & 0.0199999810911 \\
14 & 0.146153846154 & 0.0199999803259 \\
15 & 0.147540983607 & 0.0199999798308 \\
16 & 0.148760330579 & 0.0199999793944 \\
17 & 0.151260504202 & 0.0199999784968 \\
18 & 0.152542372881 & 0.0199999780356 \\
19 & 0.153225806452 & 0.0199999777895 \\
20 & 0.155172413793 & 0.0199999770888 \\
21 & 0.155737704918 & 0.0199999768851 \\
22 & 0.15652173913 & 0.0199999766031 \\
23 & 0.15873015873 & 0.0199999758093 \\
24 & 0.161016949153 & 0.01999997499 \\
25 & 0.161538461538 & 0.0199999748037 \\
26 & 0.163636363636 & 0.0199999740568 \\
27 & 0.16393442623 & 0.0199999739511 \\
28 & 0.165217391304 & 0.019999973497 \\
29 & 0.168 & 0.0199999725204 \\
30 & 0.169491525424 & 0.0199999720019 \\
31 & 0.169642857143 & 0.0199999719496 \\
32 & 0.175 & 0.019999970126 \\
33 & 0.177419354839 & 0.0199999693252 \\
34 & 0.177966101695 & 0.0199999691464 \\
35 & 0.1796875 & 0.0199999685892 \\
36 & 0.181034482759 & 0.0199999681599 \\
37 & 0.181102362205 & 0.0199999681384 \\
38 & 0.185840707965 & 0.019999966678 \\
39 & 0.188679245283 & 0.0199999658436 \\
40 & 0.188976377953 & 0.0199999657582 \\
41 & 0.189189189189 & 0.0199999656973 \\
42 & 0.191304347826 & 0.0199999651016 \\
43 & 0.192 & 0.0199999649102 \\
44 & 0.198198198198 & 0.0199999633028 \\
45 & 0.198412698413 & 0.0199999632505 \\
46 & 0.201612903226 & 0.0199999624987 \\
47 & 0.203125 & 0.019999962162 \\
48 & 0.230088495575 & 0.0199999582227 \\
49 & 0.237288135593 & 0.0199999577104 \\
50 & 0.24347826087 & 0.0199999573218 \\
51 & 0.75652173913 & 4.26781661018e-08 \\
52 & 0.762711864407 & 4.22895950203e-08 \\
53 & 0.769911504425 & 4.17773267196e-08 \\
54 & 0.796875 & 3.78379607993e-08 \\
55 & 0.798387096774 & 3.75012747522e-08 \\
56 & 0.801587301587 & 3.67495175436e-08 \\
57 & 0.801801801802 & 3.66971725473e-08 \\
58 & 0.808 & 3.50898496961e-08 \\
59 & 0.808695652174 & 3.48983825796e-08 \\
60 & 0.810810810811 & 3.43027436629e-08 \\
61 & 0.811023622047 & 3.42417803903e-08 \\
62 & 0.811320754717 & 3.41564267078e-08 \\
63 & 0.814159292035 & 3.33219847999e-08 \\
64 & 0.818897637795 & 3.18616196586e-08 \\
65 & 0.818965517241 & 3.18401194958e-08 \\
66 & 0.8203125 & 3.14107514374e-08 \\
67 & 0.822033898305 & 3.08536200685e-08 \\
68 & 0.822580645161 & 3.06748238685e-08 \\
69 & 0.825 & 2.98739995658e-08 \\
70 & 0.830357142857 & 2.80504457925e-08 \\
71 & 0.830508474576 & 2.79981454722e-08 \\
72 & 0.832 & 2.74796016497e-08 \\
73 & 0.834782608696 & 2.65030131452e-08 \\
74 & 0.83606557377 & 2.6048932282e-08 \\
75 & 0.836363636364 & 2.59431801354e-08 \\
76 & 0.838461538462 & 2.51963005113e-08 \\
77 & 0.838983050847 & 2.50100240309e-08 \\
78 & 0.84126984127 & 2.41907308872e-08 \\
79 & 0.84347826087 & 2.33969247032e-08 \\
80 & 0.844262295082 & 2.3114891539e-08 \\
81 & 0.844827586207 & 2.29112366856e-08 \\
82 & 0.846774193548 & 2.22104586632e-08 \\
83 & 0.847457627119 & 2.19643728934e-08 \\
84 & 0.848739495798 & 2.15032273376e-08 \\
85 & 0.851239669421 & 2.06056457383e-08 \\
86 & 0.852459016393 & 2.01692289814e-08 \\
87 & 0.853846153846 & 1.96740837441e-08 \\
88 & 0.856 & 1.89088358651e-08 \\
89 & 0.856 & 1.89089067043e-08 \\
90 & 0.857142857143 & 1.85048362883e-08 \\
91 & 0.860869565217 & 1.71996949912e-08 \\
92 & 0.864406779661 & 1.59815002378e-08 \\
93 & 0.864406779661 & 1.59814843581e-08 \\
94 & 0.866071428571 & 1.54164153422e-08 \\
95 & 0.873949579832 & 1.28273384345e-08 \\
96 & 0.876033057851 & 1.21695452639e-08 \\
97 & 0.880733944954 & 1.07324790583e-08 \\
98 & 0.885964912281 & 9.21731228148e-09 \\
99 & 0.88785046729 & 8.69445239068e-09 \\
100 & 0.901639344262 & 5.28563820837e-09 \\
}\optB
\begin{tikzpicture}
  \begin{axis}[
      width=\linewidth,
      height=55mm,
      y tick label style={
        /pgf/number format/.cd,
        fixed,
        fixed zerofill,
        precision=2,
        /tikz/.cd
      },
      legend style={at={(0,1)},
        anchor=north west,legend columns=-1},
      xtick={20,40,60,80,100},
      nodes near coords align={vertical},
      ymin=0,
      xmin=1,xmax=101,
      xlabel={\small{Tree index}},
      ]
    \addplot[ybar interval, ybar legend,blue,fill=blue] table[x=h,y=rho]{\optB};
    \addplot[red] table[x=h,y=risk]{\optB};
    \legend{$\posterior_C$, {$\hat{L}(h_i,\oobvalset_i)$}}
  \end{axis}
\end{tikzpicture}
  \end{center}
  \caption{Optimized distribution $\posterior_\C$ and for each tree $i$ the
    error on the subset $\oobvalset_i$  of the training data not used for
    building the tree.
    Shown are the results for the \dataset{Credit-A} data set using 100
    self-complemented trees.
  }
  \label{fig:rho-b}
\end{figure}
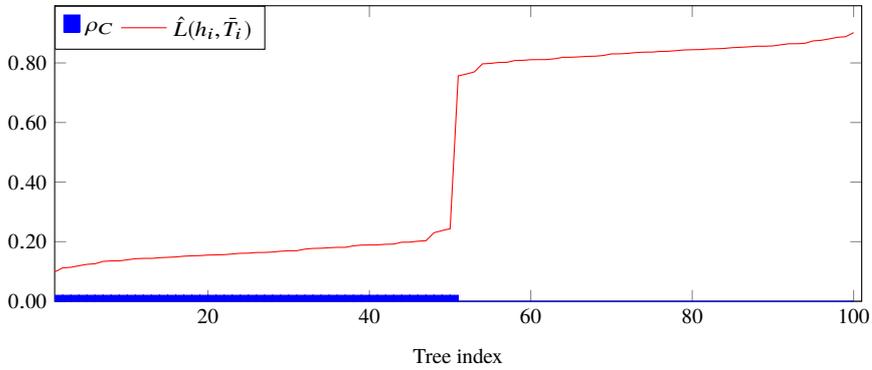
Figure~\ref{fig:rho-b} shows an example. The algorithm selected almost exclusively the strong classifiers, and due to the required alignment, the $\posterior_\C$ was basically the uniform distribution on the original (non-self-complemented) hypothesis set, explaining the similarities in accuracy and bounds.

\section{Conclusions}
\label{section:conclusion}
PAC-Bayesian generalization bounds can be used to obtain rigorous performance guarantees for the standard random forest classifier used in practice.
No modification of the algorithm is necessary and
no additional data {is} required because the out-of-bag samples can be exploited.
In our experiments {using the standard random forest}, bounds inherited from the corresponding Gibbs classifiers
clearly outperformed  majority vote bounds that take correlations between ensemble members into account.
The reason is that the individual decision trees are already rather accurate classifiers, which makes it difficult to estimate the correlations of errors.
{As expected, we could observe the opposite result when using weaker
  individual classifiers.
However, this required enough  disagreement between the classifiers
(which we enforced by increasing randomization) and 
using a separate validation set,
because the out-of-bag samples alone
provided not enough data for reliably  estimating the correlation between two voters.}
We {also} replaced the majority vote by a weighted majority vote and 
optimized the weights by minimizing the PAC-Bayesian bounds. This led to better performance guarantees, but weaker empirical performance.

When we split the data available at training time into a  training
  and a validation set, we can use the hold-out validation set to
  compute a generalization bound. In our experiments, this led to
  considerably tighter bounds compared to the PAC-Bayesian
  approaches.
  However, because less data was available for training, the resulting
  classifiers performed worse on an external test set in most cases.
  Thus, using a validation set gave us better performance guarantees,
  but worse performance.

  Our conclusion is that existing results that are derived for
  ensemble methods and take correlations of predictions into account
  are not sufficiently strong for guiding model selection and/or
  weighting of ensemble members in majority voting of powerful
  classifiers, such as decision trees. While the \cbound s are
  empirically outperformed by the generalization bounds based on the
  Gibbs classifier, the latter ignore the effect of cancellation of
  errors in majority voting and, thus, are of limited use for
  optimizing a weighting of the ensemble members and guiding model
  selection. Therefore, more work is required for tightening the
  analysis of the effect of correlations in majority voting.
  Nevertheless, to our knowledge, the PAC-Bayesian approach in this
  study provides the tightest upper bounds for the performance of the
  canonical random forest algorithm without requiring hold-out data.

\section*{Acknowledgements}

We acknowledge support by the Innovation Fund Denmark through the \emph{Danish Center
for Big Data Analytics Driven Innovation} (DABAI).

\bibliography{ecml}
\bibliographystyle{spbasic}      

\newpage
\appendix

\section{Extra Results for Second Setting}
Table~\ref{tab:results-bagtest-no-oob} lists the bounds and losses obtained in the validation set setting using only $\valset$ for computing the bounds.

\begin{table}[h!]
\caption{
The \pbbound, \cboundempA, \cboundempB\ and \shbound\ computed for the binary UCI data sets in the validation set setting, where only $\valset$ is used for computing the bounds. The majority vote loss on $\testset$
      is given as an estimate of the accuracy of the trained
      classifier denoted as \emph{test score}.
The best bound is marked with \textbf{bold}, while \textit{italics} is used to indicate trivial bounds ($\geq 0.5$).
}
\label{tab:results-bagtest-no-oob}
\vskip 0.15in
\begin{center}
\begin{small}
\begin{sc}
\begin{tabular}{lcc|ccccc}
  \toprule
  Data set & $n$ & $d$ & Test score & \PB & $\C1$ & $\C2$ & \sh \\
  \midrule
  Adult & 45222 & 14 & 0.154 & 0.432 & \textit{0.510} & 0.483 & \textbf{0.169}\\
Credit-A & 653 & 15 & 0.135 & \textit{0.632} & \textit{0.854} & \textit{0.812} & \textbf{0.294}\\
Haberman & 306 & 3 & 0.333 & $>$\textit{1} & $>$\textit{1} & $>$\textit{1} & \textbf{\textit{0.577}}\\
Heart & 297 & 13 & 0.282 & \textit{0.893} & \textit{0.992} & \textit{0.990} & \textbf{0.412}\\
ILPD & 579 & 10 & 0.307 & $>$\textit{1} & $>$\textit{1} & $>$\textit{1} & \textbf{0.441}\\
Ionosphere & 351 & 34 & 0.125 & \textit{0.728} & \textit{0.931} & \textit{0.910} & \textbf{0.299}\\
Letter:AB & 20000 & 16 & 0.015 & 0.152 & 0.266 & 0.192 & \textbf{0.035}\\
Letter:DO & 20000 & 16 & 0.067 & 0.237 & 0.383 & 0.300 & \textbf{0.072}\\
Letter:OQ & 20000 & 16 & 0.059 & 0.352 & \textit{0.519} & 0.418 & \textbf{0.119}\\
Mushroom & 8124 & 22 & 0.001 & 0.017 & 0.036 & 0.041 & \textbf{0.008}\\
Sonar & 208 & 60 & 0.250 & $>$\textit{1} & $>$\textit{1} & $>$\textit{1} & \textbf{\textit{0.510}}\\
Tic-Tac-Toe & 958 & 9 & 0.142 & \textit{0.765} & \textit{0.908} & \textit{0.807} & \textbf{0.221}\\
USvotes & 232 & 16 & 0.052 & \textit{0.513} & \textit{0.784} & \textit{0.739} & \textbf{0.228}\\
WDBC & 569 & 30 & 0.063 & 0.342 & \textit{0.567} & 0.490 & \textbf{0.102}\\

  \bottomrule
\end{tabular}
\end{sc}
\end{small}


\end{center}
\vskip -0.1in
\end{table}

\section{Comparison Plots for the Bagging and Validation Set Settings}\label{app:comparison}
\begin{figure}[h!]
  \centerline{
    \pgfplotstableread[row sep=\\,col sep=&]{
val       & bag & bagtest & test \\
test_risk & 0.0449293966624 & 0.0667522464698 & 0.0667522464698 \\
seeger    & 0.215675647391 & 0.226750570856 & 0.236953005162 \\
c1        & 0.391344172074 & 0.362280465302 & 0.383377740246 \\
c2        & 0.530492142568 & 0.294210647719 & 0.299928423509 \\
sh        &  & 0.0720047901067 & 0.0720047901067 \\
}\compareletterDO
\begin{tikzpicture}
  \begin{axis}[
        width=0.54\linewidth,
      y tick label style={
        /pgf/number format/.cd,
        fixed,
        fixed zerofill,
        precision=2,
        /tikz/.cd
      },
      ybar,
      bar width=.2cm,
      legend style={at={(0.65,1)},
        anchor=north east,legend columns=-1},
      symbolic x coords={test_risk, seeger, c1, c2, sh},
      xticklabels={{$\hat{L}^{\text{MV}}$}, \PB, $\C1$, $\C2$, \sinhyp},
      xtick={test_risk, seeger, c1, c2, sh},
      nodes near coords align={vertical},
      ymin=0,ymax=0.7,
    ]
    \addplot[blue,fill=blue] table[x=val,y=bag]{\compareletterDO};
    \addplot[green,fill=green] table[x=val,y=test]{\compareletterDO};
    \addplot[red,fill=red] table[x=val,y=bagtest]{\compareletterDO};
    \legend{$\oobvalset$, $\valset$, $\oobvalset$+$\valset$}
  \end{axis}
\end{tikzpicture}
    \hfill
    \input{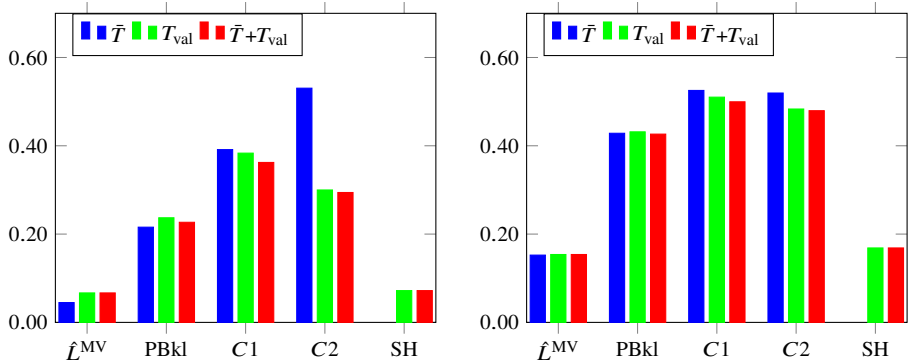}
\begin{tikzpicture}
  \begin{axis}[
        width=0.54\linewidth,
      y tick label style={
        /pgf/number format/.cd,
        fixed,
        fixed zerofill,
        precision=2,
        /tikz/.cd
      },
      ybar,
      bar width=.2cm,
      legend style={at={(0.65,1)},
        anchor=north east,legend columns=-1},
      symbolic x coords={test_risk, seeger, c1, c2, sh},
      xticklabels={{$\hat{L}^{\text{MV}}$}, \PB, $\C1$, $\C2$, \sinhyp},
      xtick={test_risk, seeger, c1, c2, sh},
      nodes near coords align={vertical},
      ymin=0,ymax=0.7,
    ]
    \addplot[blue,fill=blue] table[x=val,y=bag]{\compareadult};
    \addplot[green,fill=green] table[x=val,y=test]{\compareadult};
    \addplot[red,fill=red] table[x=val,y=bagtest]{\compareadult};
    \legend{$\oobvalset$, $\valset$, $\oobvalset$+$\valset$}
  \end{axis}
\end{tikzpicture}
  }
\caption{Comparison of the bounds obtained for a random forest with
  500 trees trained on the \dataset{Letter:DO} data set. Comparison of the bounds obtained for a random forest with 500 trees trained on the \dataset{Letter:DO} data set (left) and for a random forest with 1000 trees trained on the \dataset{Adult} data set (right).}
\label{fig:comparison}
\end{figure}

Figure~\ref{fig:comparison} shows the comparison of the bounds obtained for the \dataset{Letter:DO} and \dataset{Adult} data set. The figure includes all three settings: using only the hold-out sets from bagging ($\oobvalset$), using only the validation set ($\valset$), and using a combination of both ($\oobvalset$+$\valset$)

\end{document}